\documentclass{vgtc}                          

\graphicspath{{figures/}{pictures/}{images/}{./}} 

\usepackage{times}       
\usepackage{afterpage}

\usepackage{tabu}                      
\usepackage{booktabs}                  
\usepackage{lipsum}                    
\usepackage{mwe}                       

\usepackage{mathptmx}                  

\onlineid{0}

\vgtccategory{Research}

\vgtcinsertpkg

\title{Visualization Literacy of Multimodal Large Language Models: \\ A Comparative Study}

\author{Zhimin Li\thanks{e-mail: zhimin@sci.utah.edu}\\ %
        \scriptsize University of Utah %
\and Haichao Miao\thanks{e-mail: miao1@llnl.gov}\\ %
     \parbox{1.4in}{\scriptsize \centering Lawrence Livermore National Laboratory} %
\and Valerio Pascucci \thanks{e-mail: pascucci@sci.utah.edu}\\ %
     \parbox{1.4in}{\scriptsize \centering University of Utah}
\and  Shusen Liu \thanks{e-mail: liu42@llnl.gov}\\ %
     \parbox{1.4in}{\scriptsize \centering Lawrence Livermore National Laboratory}}

\abstract{The recent introduction of multimodal large language models (MLLMs) 
combine the inherent power of large language models (LLMs) with the renewed capabilities to reason about the multimodal context. The potential usage scenarios for MLLMs significantly outpace their text-only counterparts. 
Many recent works in visualization have demonstrated MLLMs' capability to understand and interpret visualization results and explain the content of the visualization to users in natural language.
In the machine learning community, the general vision capabilities of MLLMs have been evaluated and tested through various visual understanding benchmarks.
However, the ability of MLLMs to accomplish specific visualization tasks based on visual perception has not been properly explored and evaluated, particularly, from a visualization-centric perspective. 

In this work, we aim to fill the gap by utilizing the concept of visualization literacy to evaluate MLLMs. 
We assess MLLMs' performance over two popular visualization literacy evaluation datasets (VLAT and mini-VLAT).
Under the framework of visualization literacy, we develop a general setup to compare different multimodal large language models (e.g., GPT4-o, Claude 3 Opus, Gemini 1.5 Pro) as well as against existing human baselines. 
Our study demonstrates MLLMs' competitive performance in visualization literacy, where they outperform humans in certain tasks such as identifying correlations, clusters, and hierarchical structures.}


\begin{document}


\maketitle
\section{Introduction}

Recently, due to the advance of generative neural network models such as large language models (LLMs), we are able to easily produce visualization based on natural language instruction by generating visualization code. 
However, due to the modality of model input, LLMs can not easily understand the generated visualization in image form. 
That limitation changes with the introduction of multi-modal capability, i.e., multi-modal large language models (MLLMs), in which the LLMs also obtain the ability to understand vision input, including visualization results \cite{yang2023dawn}. 
Such capabilities facilitate novel usage of MLLMs, e.g., developing agents that can accomplish potentially complex visualization tasks \cite{liu2023ava}.

However, all these potential use cases hinged on the model's ability to understand the visualization, which we know little about, particularly, regarding the extent of their ability and limitations. 
As a result, a systematic evaluation of the ability of MLLMs to understand the visualization is the critical step we need before applying them in downstream applications.
Several benchmarks have been introduced in the machine learning community for evaluating the model's capabilities for chart and infographic understanding 
\cite{masry2022chartqa, mathew2022infographicvqa, kahou2017figureqa}. They provide valuable assessment regarding the general capability for understanding information in graphics form, however, due to their unstructured nature, it is hard to understand the model's performance on specific types of visual encoding or visualization tasks. Moreover, these datasets are not designed with visualization evaluation in mind, therefore the example selection process is likely non-discriminative in terms of poor visualization design or includes many illustrations instead of proper visualization.

In this work, we aim to fill this gap by obtaining a fine-grain assessment of MLLM's behavior on a smaller yet well-curated dataset that covers major visual encoding and visualization tasks. We achieve this by bringing the concept of visualization literacy and the associated tests for evaluating the MLLM's capability for understanding visualization. We compared multiple state-of-the-art models, as well as against human performance from previous visualization literacy research on the same test. 

Literacy often refers to the ability to read and write. In a broader sense, it encompasses the capacity to effectively understand, interpret, and communicate non-verbal information in written form. Following a similar definition, visualization researchers brought the concept of visualization literacy, which describes a person's ability to interpret visual communication through visualizations such as various charts and graphs. A curated set of questions has been proposed to cover different types of visual encoding with varying types of questions. Despite the small size of these questions and examples, they aim for a broad coverage both in terms of visual encoding and the type of queries. Most notably, VLAT and Mini-VLAT, present well-recognized and well-established datasets~\cite{nobre2024reading,pandey2023mini,lee2016vlat} for visualization literacy assessment.

In this work, we investigate how well MLLM models can perform on these visualization literacy tests. 
More importantly, what kind of mistakes that the LLM is prone to make, and what are their strength and weakness?
Through our study, we find that MLLMs are good at visualization tasks such as correlation trends and cluster analysis. They also demonstrate an outstanding performance on treemap interpretation. 
Despite the competitive performance of MLLMs over visualization literacy, we observe that humans and MLLMs demonstrate distinct failure patterns.

Our key contributions are summarized below:
\begin{itemize}
    \item Introduce the concept of visualization literacy for evaluating MLLMs' capabilities for understanding visualization.
    
    \item Present a comparative evaluation of multiple stat-of-art MLLMs over the Mini-VLAT and VLAT visualization literacy test, as well as their performance against an open-source human evaluation result on the same test.
    
    \item Provide the accompanying error analysis for uncovering the limitation of current MLLMs for visualization tasks and encodings. 
\end{itemize}

\section{Related Work}

\subsection{Visualization competency test}
A person's ability to understand visualization and visual encoding can vary greatly depending on their education level and prior exposure. 
Börner et al.~\cite{borner2016investigating} evaluate the 273 museum visitors with different backgrounds including youth and adults. The evaluation visualization is 20 visualizations that show up in newspapers and textbooks. They find that many interview candidates cannot understand and interpret the presented visualization. The observation implies information visualization is not well-accessed by the population. 
Various visualization competency tests have been introduced as part of visualization education efforts.
Alper et al.~\cite{alper2017visualization} explored how visualization is taught at elementary school and interviewed the school teacher about the challenge of visualization teaching. A web application is developed at the end to improve the visualization literacy of children. 
E Firat~\cite{firat2020treemap} develop the treemap literacy which specifically focuses on evaluating the human's performance over treemap visualization.
VLAT~\cite{lee2016vlat} is the well-established test used to measure visualization literacy. The overall evaluation contains 12 visualizations and 53 questions. Pandey and Ottley~\cite{pandey2023mini} shorted the test into a Mini-VLAT dataset which only contains 12 questions. Carolina et al.~\cite{nobre2024reading} replicate the evaluation study of the VLAT evaluation and give a more detailed understanding of the barriers during the visualization interpretation process.
The discussion of the previous visualization competency research mainly assesses humans' capabilities. However, in this work, the main focus of our study is MLLM. We compare their performance with each other and reason about their mistakes. 

\subsection{MLLM and LLM for Visualization}
The MLLMs and LLM have demonstrated outstanding abilities to understand and solve a wide range of practical problems \cite{yang2023dawn}.
Many research efforts have utilized the generative capability of large language model ~\cite{vazquez2024llms, Xu2024ExploringTC, li2024visualization} to create visualization through code generation or to interpret and manipulate the visualization with the SVG format.
Chen et al.~\cite{chen2023beyond} evaluate the performance of GPT3.5 and GPT4 over a data visualization course and find out that the LLM can score 80\% of the assignment. Such a finding motivates the new requirement of visualization education design in college. 
In rich screen reader, Zong et al.~\cite{zong2022rich} use LLM to describe visualization to people with impaired vision.
Yang et al.~\cite{yang2024foundation} present a review study of how the foundational model will help visualization over the previous literature.
%
Liu. et al.~\cite{liu2023ava} proposed the concept of the autonomous visualization agent (AVA), which utilizes the visual perception ability of MLLMs to directly understand user intention adjust visualization parameters, and help solve visualization tasks autonomously. 
Compared with our study, most of these works do not involve the new emerging perception ability of MLLMs, and they often only focus on models from a single source (e.g., OpenAI).
Instead, our study focuses on the vision aspect of MLLMs and evaluates their visualization literacy. 

\subsection{Machine perception for visualization}
Visual perception is a fundamental piece of the visualization system design. Due to the complexity of human evaluation, many researchers aim to use machine learning models (e.g., neural networks) to assist or even replace human evaluation. Convolution neural network is a potential candidate for such a task given that its initial design is inspired by the mechanism of humans' visual cortex~\cite{lecun1995convolutional}.
Giovannangeli, Loann, et al~\cite{giovannangeli2020toward} use this idea to predict the performance of node-link diagrams and adjacency matrix diagrams over humans with a neural network evaluation. 
Haehn et al.~\cite{haehn2018evaluating} study the analysis ability of  convolution neural networks on visualization tasks by evaluating their performance over elementary perceptual tasks from Cleveland and
McGill~\cite{cleveland1984graphical}. Their study concludes that CNN is not an ideal model for human perception modeling which is similar to previous discovering~\cite{bowers2023deep}.
Yan et al.~\cite{yang2023can} study over twenty-nine neural network architectures to predict human behavior over the correlation comparison in scatter plots.
There are more works~\cite{giovannangeli2020toward,zhang2020viscode} under this path to study the correlation between neural networks and human performance.
Compared with previous work, our study focuses on more powerful Multimodal large language models that are pre-trained with significantly larger datasets than the traditional network models (e.g., CNN pretrained with imageNet data). 
Moreover, our evaluation focuses on understanding MLLMs' ability in visualization literacy in general instead of a specific visualization or task.



\section{Background}

\subsection{Multimodal Vision Language Models}
Recently, models that accept multiple modalities as input are gaining increasing popularity. From the early work, e.g., CLIP~\cite{radford2021learning}, that jointly embedded image and text to models designed for input in one modality and output in another, such as text-to-image \cite{rombach2022high, ramesh2022hierarchical} and image-to-text \cite{yu2022coca} models.
In the context of this work, we focus on multi-modal large language models (LLM) \cite{yin2023survey} that can understand both image and text inputs for text generation.
LLMs are often referred to as foundation models due to their extensive capabilities and scale. They often possess a wide range of knowledge, allowing them to understand implicit context and common sense. 
Since humans interact with their environment through multi-modal sensory input, the evolution from text-only LLM systems to multi-modal LLMs significantly increases the potential usage cases for LLMs, prompting many applications \cite{yang2023dawn}. Since the initial introduction of GPT4-V, many state-of-the-art MLLMs have been introduced such as the recent update to GPT4-o from OpenAI, multimodal support for Claude 3 \cite{anthropic2024claude3} from Anthropic, and Gemini family from Google. On the open weight model side, many attempts have been made to develop similar capabilities, such as LLAVA \cite{liu2023llava}, and CogVLM \cite{wang2023cogvlm}.

\subsection{Visualization Literacy}
In the Merriam-Webster dictionary, literacy is defined as ``the ability to read and write''
Visualization literacy is often described as the ability to understand and extract information from data visualization. Understanding visualization literacy has attracted a broad attention~\cite{lee2016vlat, boy2014principled, alper2017visualization, borner2016investigating}.
In particular, VLAT data is a popular and well-established dataset that is used to measure visualization literacy~\cite{pandey2023mini, nobre2024reading} of humans.
The dataset contains 12 different visualizations with 53 questions. In this study, we collect our visualization literacy test from the original authors' website \footnote{\href{https://www.bckwon.com/publication/vlat/}{https://www.bckwon.com/publication/vlat/}}.

Mini-VLAT is a short version of the VLAT dataset which has the same 12 visualizations but only presents 12 questions. We collect this dataset from the original authors' repository\footnote{\href{https://github.com/washuvis/minivlat/blob/main/MiniVlatQUESTIONS.pdf}{https://github.com/washuvis/minivlat/blob/main/MiniVlatQUESTIONS.pdf}}. 
Previous research~\cite{pandey2023mini} shows that the behavior of humans in the Mini-VLAT dataset shows similar behavior patterns over the VLAT dataset.
We use the open-source human evaluation result \footnote{\href{https://osf.io/dy67k/}{https://osf.io/dy67k/}}. This dataset is collected by the author from the Mini-VLAT to replicate the previous VLAT study. The total number of people who finish the test is 199 participants. Most of them have high school diplomas. 19\% have a Master's degree and around 40\% have a bachelor's degree.
During the question and answer process, each question needs to be completed in 25 seconds.
The author claims that the overall score of this replicate study is similar to the original VLAT study~\cite{lee2016vlat}. More detail about the study can be found in the original experiment~\cite{pandey2023mini}.

\section{Experiment Design}
To perform a meaningful comparison, we choose the best MLLMs from the frontier labs, namely, GPT4-o, Gemini Pro, and Claude 3 Opus, as the candidate models and evaluate their performance.
They are the most powerful models from the most dominant companies in the LLM space and often demonstrate state-of-the-art performance over different evaluation benchmarks in the current literature.

The experiment tests each multimodal language model on the Mini-VLAT and VLAT datasets.
It is important to notice that LLM can give different answers to the same question in separate runs (due to the random seed). 
Therefore, each LLM will be tested 10 times on the same question separately.
The final result for each question is reported as the average score of all 10 runs (each correct answer gets a score of 1.0, whereas the incorrect answer gets a score of 0.0).
The overall testing and scoring process is fully automatic and the model query is through Python script which is implemented with the API of OpenAI, Google, and Anthropic.
For LLM such as GPT4-o, it costs $0.26\$$ and less than 5 minutes to answer all 53 questions from the VLAT dataset. As a reference, hiring a human to perform the same task~\cite{nobre2024reading} takes around a half hour to finish the test.

How to ask a question can affect the quality of its answer.
Similarly, how to ask an LLM question, which is often considered as the prompt, can significantly affect the performance of an LLM~\cite{wei2022chain}. 
In our setup, we aim to use a simple and functional prompt that introduces as little interference or perturbation to the original model's capability as possible. Optimizing prompts can be an extensive undertaking and can also lead to implicit bias toward one model or another. As a result, in our experiment, we did not optimize the prompt or employ enhanced reasoning approaches such as chain-of-thoughts \cite{wei2022chain}. 
We use the same prompt for all models with the following format.
An example prompt for one question given a visualization:\\ \\
\textit{What was the price of a barrel of oil in February 2015?
Answer: 1) \$57.36 2) \$47.82 3) \$50.24 4) \$39.72.\\
Please answer with Answer: 1) - 4) and Why:}
\\
\\
Beyond the question and answer, the response format helps us produce more consistent answers for automatic scoring. The ``why'' is included to provide a bit more context for the answer for our subsequent analysis. 





\section{Evaluation}

\subsection{Evaluation on Mini-VLAT}

\begin{figure}[t]
\centering   
    \includegraphics[width=\linewidth]{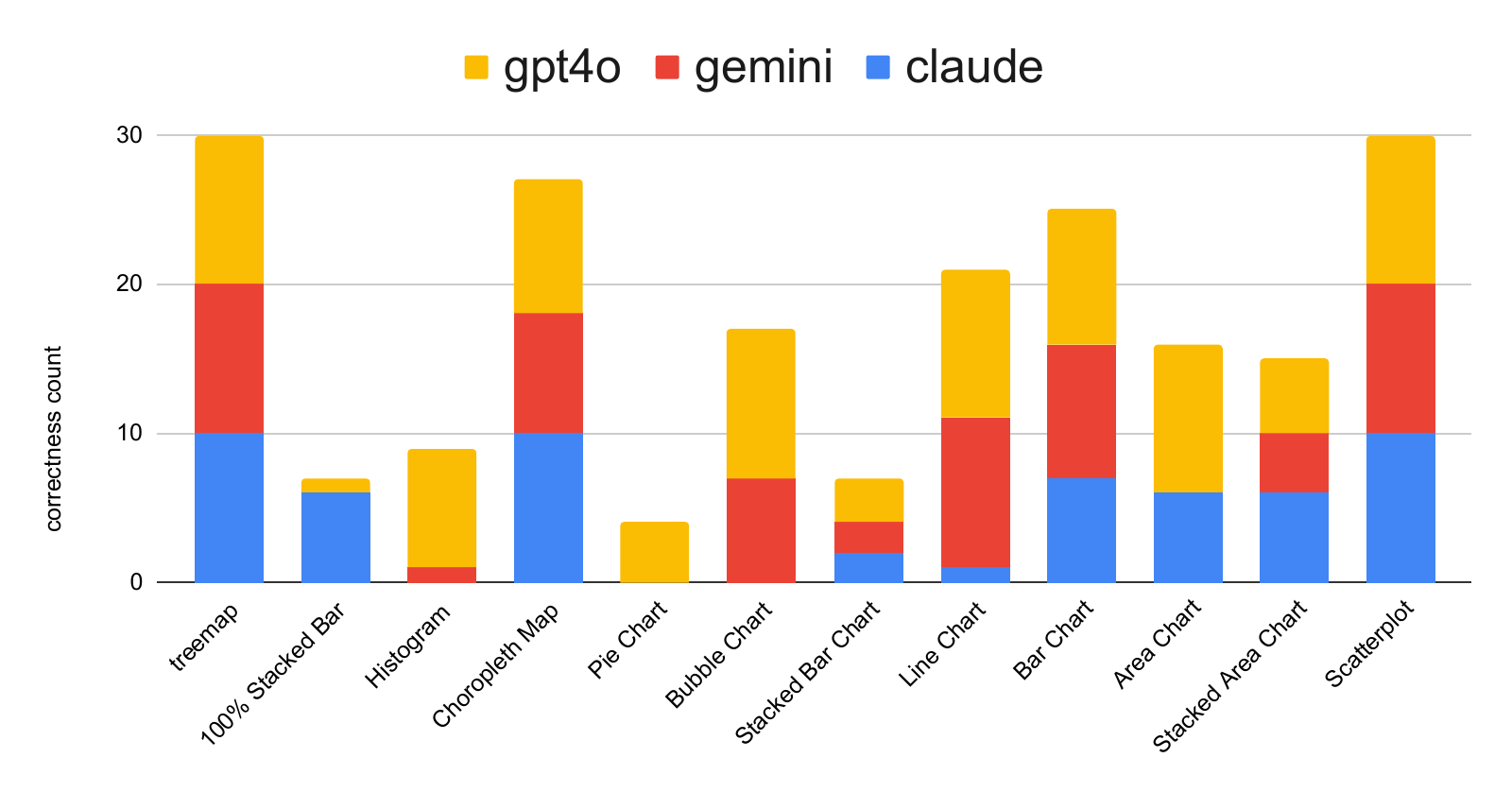}
    \caption{The result shows the performance of different multimodal large language models over the Mini-VLAT dataset.}
\label{fig:mini_vlat_question_result}
\end{figure}

\begin{table}[t]
\centering
\begin{tabular}{cccccccc}
 \hline
MLLM   & GPT4-o & Claude-opus & Gemini-1.5-pro & RB\\  \hline
 best    & \textbf{0.833} & 0.75 &  0.666 &  --\\ 
 worst    & \textbf{0.583} & 0.333 & 0.333   &  --\\ 
 mean  & \textbf{0.741 } & 0.483 & 0.508 &  0.35\\ 
  \hline
\end{tabular}
\caption{The performance of different MLLMs over the Mini-VLAT datasets.}
\label{tab:mini_vlat_data_table}
\end{table}

We perform a pilot evaluation study with the Mini-VLAT dataset to compare three state-of-the-art MLLMs' performance alongside a random selection baseline to help gauge the models' performance (the questions are with different numbers of choices, so the score of random guesses is not obvious).
The summary of the result is described in TABLE~\ref{tab:mini_vlat_data_table}. 
The best/worst performance measures the highest/lowest single run score out of all runs. The average score is the mean of all runs.
The best and average performance of each model is significantly better than the random baseline.
In this case, the GPT4-o, which was recently released by OpenAI, has the best statistic over all three measurements.
Gemini has the second performance on average, while the Claude model has the advantage in terms of its best run.
The detailed performance breakdown of each question is described in Fig.~\ref{fig:mini_vlat_question_result}. All models correctly answer questions for the treemap and scatter plot visualization. They also perform well over the Choropleth chart \cite{andrienko2001choropleth}, Line chart, and Bar chart. The GPT4-o model performs better than others in the Histogram chart and Area chart.

\subsection{Error Analysis}
To better understand the model's behavior, it is crucial to investigate beyond just summary statistics and understand where and how the model fails.
Existing literature has discussed the failure reasons from the human visualization literacy tests~\cite{nobre2024reading}. In this section, we will dive into the details and try to understand why an MLLM fails a question. Our discussion focuses on the question in the Mini-VLAT dataset. 

Given the overall performance of GPT4-o outperforms other models, we mainly discuss the error analysis of the GPT4-o model. Specifically, our analysis focuses on the 100\% Stacked Bar chart, Pie Chart, and Stacked Bart Chart.
The success rate of GPT4-o over the Stack Area Chart is also low. However, this question is related to the value-retrieving task which is the same as the Pie Chart and Stacked Bar Chart. So, we will not discuss the details here.
During the discussion, we hypothesize the reason for failure and try to verify the hypothesis by adding annotation in an attempt to remove potential confusion and improve the success rate. This process will iterate 2 to 3 times until the MLLM can pass all the question tests.

\subsubsection{Confusion from Color-Semantic Association}
\begin{figure}[t]
\centering   
    \includegraphics[width=\linewidth]{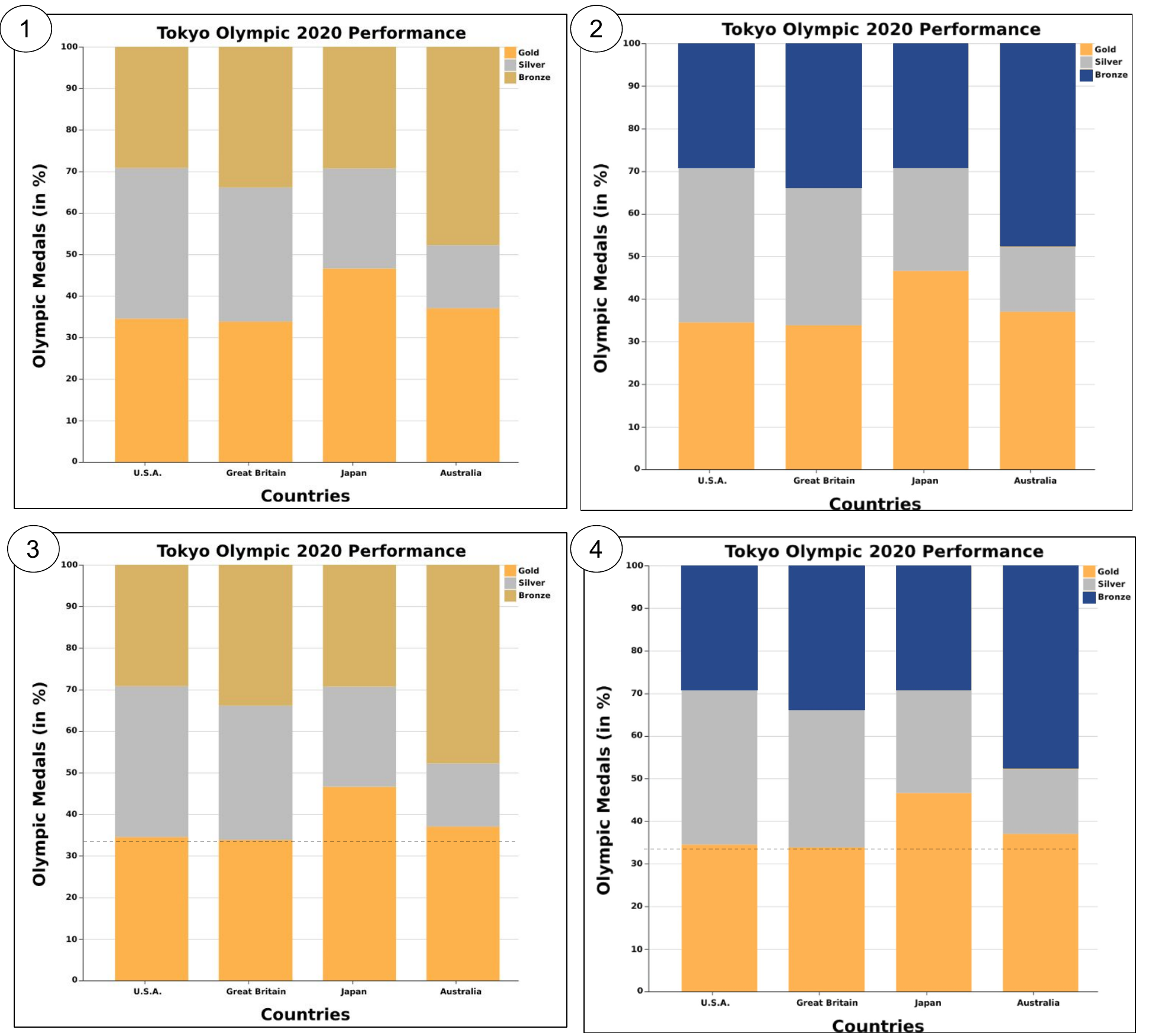}
    \caption{The language model may confused by the ambiguous colors and resort to their prior knowledge root in color-semantic association \cite{mukherjee2024estimating}, which leads to incorrect answers. }
\label{fig:vlat_question_color_problem}
\end{figure}

The color chosen to encode the visualization may affect the decision of the LLM model. 
Fig.~\ref{fig:vlat_question_color_problem} displays a 100\% stack bar chart, and the question is:

\textit{\textbf{Ask:} “Question: Which country has the lowest proportion of Gold medals?
Answers: 1) Great Britain, 2) USA, 3) Japan, 4) Australia. Please answer with Answer: 1) - 4) and Why:”}

\textit{\textbf{Response}: "Answer: 3) Japan Why: The chart shows that Japan has the lowest proportion of Gold medals compared to the other countries, as indicated by the smallest relative height of the gold segment in the stacked bar for Japan."}

For this question, Japan and Australia are frequent answers and each occurs rate is 4 out of 10 run, i.e., $4/10$. The rest of the answers are $1/10$ U.S.A and $1/10$ Great Britain. 
Here, Great Britain is the correct option.
However, Japan has the highest percentage of golden metal, and the answer from MLLM does not seem to be reasonable.
In the visualization (fig.~\ref{fig:vlat_question_color_problem} (1)), the brownish golden color and the orange color not only look similar, but the color indicates the brown metal is more likely to be referred to as golden color compared to the orange that represents the golden metal in the plot. From the explanation of the answer, we can tell that the MLLM model considers the brownish golden color as the golden percentage and considers Japan as the final answer. This explanation shows that MLLM is confused about what color should represent the golden metals, and potentially the existing color-semantic association may lead to such a mistake, a similar phenomenon has been observed in MLLMs~\cite{mukherjee2024estimating}.

To verify the hypothesis that MLLM may be confused with golden color, we updated the visualization in two different ways.
In Fig.~\ref{fig:vlat_question_color_problem} (3), we add a dashed line to help the MLLMs to compare the golden metals over different countries.
With the assistance of the dashed line, the response score of the MLLM is  5/10 Japan and 5/10  Great Britain.
The success rate of the model is improved but the MLLM is still confused with the color, and Japan is still a frequent answer in the updated visualization.
In Fig.~\ref{fig:vlat_question_color_problem} (2), we updated the color to dark blue and performed the same experiments 10 times to measure the overall success rate. 
The final statistic is 7/10 Australia and 3/10 Great Britain.
The final answer is significantly shifted with the new visualization.
In the visualization, the differences between Great Britain and Australia are still minor which is a potential reason why the MLLM failed the task in the new visualization.
In the end, we apply both updates in the visualization in Fig.~\ref{fig:vlat_question_color_problem} (4) and experiment again. The response is  10/10 Great Britain.
This experiment helps reveal the potential reason that an MLLM fails a question here is the color and the minor difference between the two categories during comparison. 

\subsubsection{Poor Performance at Value Retrieval}
\begin{figure}[t]
\centering   
    \includegraphics[width=\linewidth]{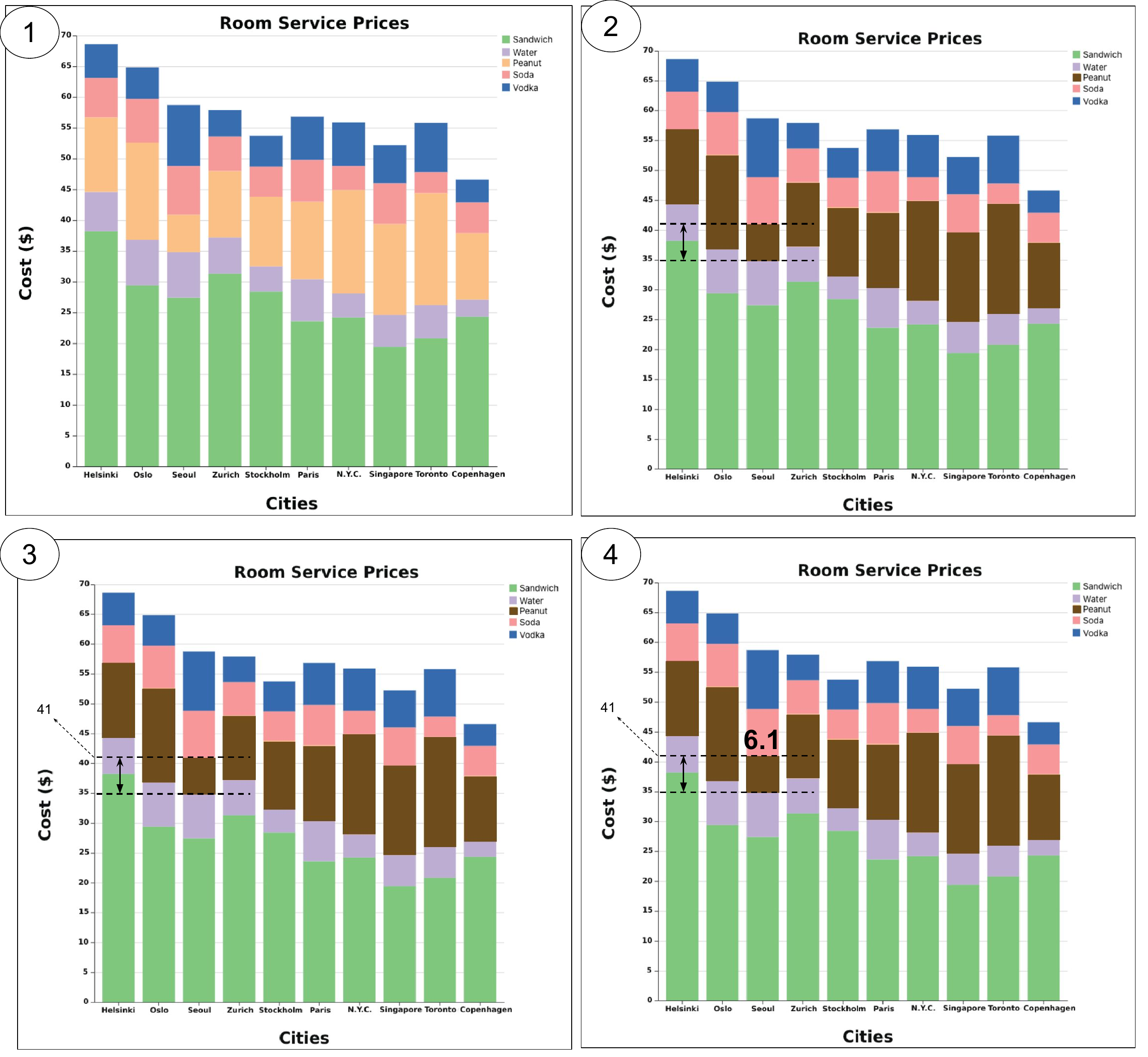}
    \caption{The language model has limited ability to retrieve value.}
\label{fig:vlat_question_insensitive_problem}
\end{figure}

Furthermore, we investigate the relatively poor performance for value retrieval tasks, here we take the Stacked Bar Chart as an example. The task is to measure the length of a rectangle. 
In Fig.~\ref{fig:vlat_question_insensitive_problem} (1), the distribution of the answer out of 10 runs is as follows, 
5/10 the answer is 7.5, 3/10 is 6.1 , and 2/10 is 5.2.

In Fig.~\ref{fig:vlat_question_insensitive_problem} (2), during the exploration and error analysis, we worried the lack of contrasts between colors may cause such a mistake during the decision process. We updated the visualization color of the peanut to make the overall comparison more distinguishable, but the final answer shows that 9/10 is 7.5, and 1/10 is 6.1.
The answers from the MLLM become more stable but the performance decreases. So likely, the color is not the main contributor to the mistakes.
Now, let's take a closer look at the question and response.

\textit{\textbf{Ask:} “Question: What is the cost of peanuts in Seoul?
Answers: 1) 5.2, 2) 6.1, 3) 7.5, 4) 4.5. Please answer with Answer: 1) - 4) and Why:”}

\textit{\textbf{Response}: “Answer: 3) 7.5 Why: To find the cost of peanuts in Seoul, we need to look at the height of the brown section of the stacked bar for Seoul. The cost for each item can be deduced by noting the cumulative height of the sections up to the top of the brown segment.
The green section (Sandwich) ends at about 30.
The purple section (Water) ends at about 35.
The brown section (Peanut) ends at approximately 42.5.
The height of the brown section (Peanut) can be calculated as the difference between the cumulative heights of the brown and purple sections: 42.5 - 35 = 7.5.
Therefore, the cost of peanuts in Seoul is 7.5. ”}

From the response of GPT4-o, we can tell that the MLLMs consider the top rectangle to be aligned with 42.5. However, the axis ticks tell us that the top rectangle has not reached the half region between 40 and 45, and the aligned value should be smaller than 42.5.
The GPT4-o fails to retrieve a correct value to do the analysis.
To help GPT4-o better understand this visualization, in Fig.~\ref{fig:vlat_question_insensitive_problem} (3), we add additional annotations (labeling the correct axis value) in the visualization to help MLLMs understand the visualization and retrieve the value.
With the assistance of the new visualization, the response of the MLLMs changed to 7.5 (5/10 runs), , and 6.1 (5/10 runs).
We consider the question in Fig.~\ref{fig:vlat_question_insensitive_problem} (3) to be simplified but the model still demonstrates a limited ability to answer this question correctly. 
In the end, we have to do the math and add the measurement result to the visualization (Fig.~\ref{fig:vlat_question_insensitive_problem} (4)) to help the MLLM pass all the tests.

\subsubsection{Difficulty in Interpreting Pie-Chart Visual Encoding}
\begin{figure}[t]
\centering   
\includegraphics[width=\linewidth]{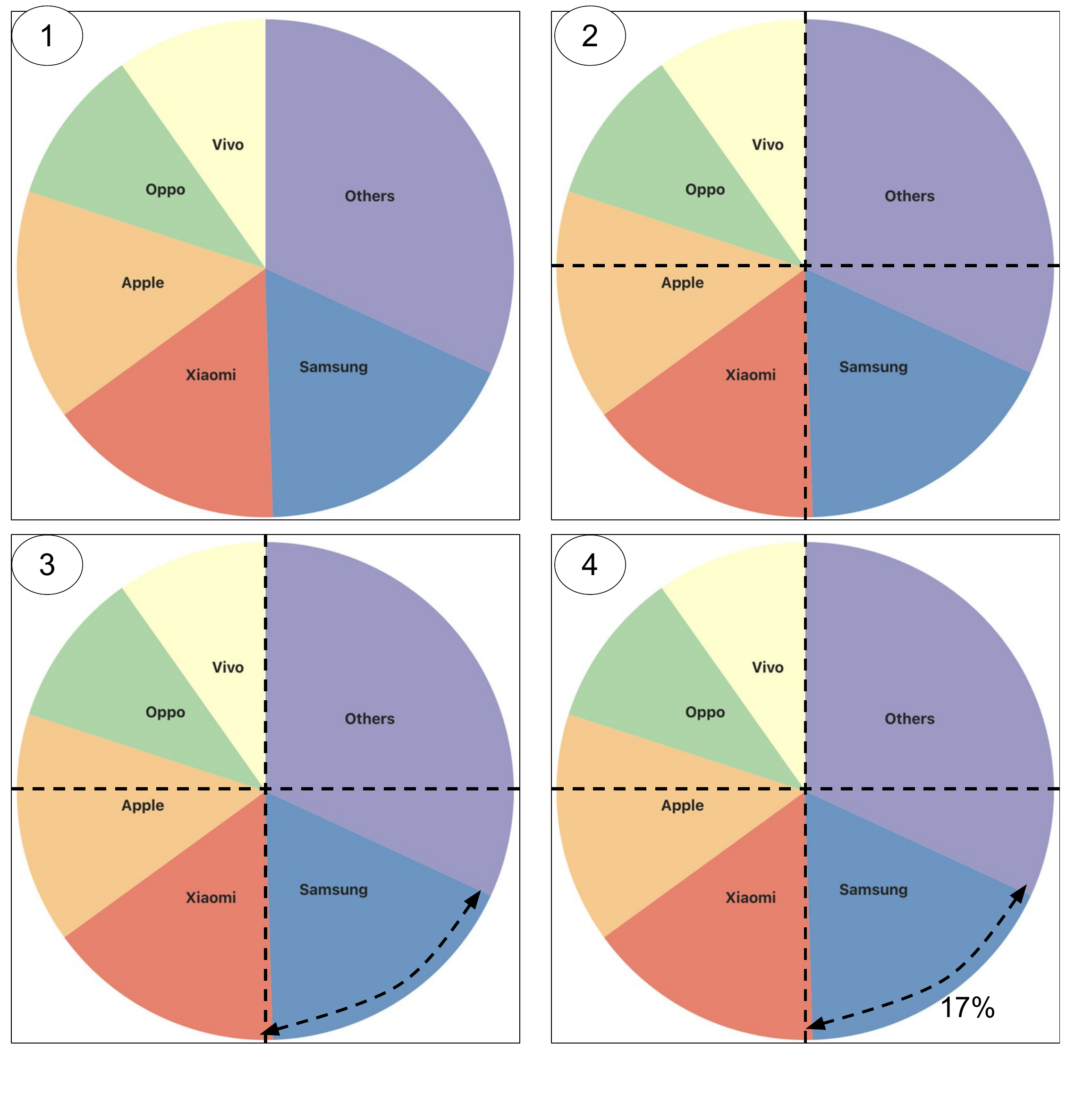}
    \caption{The language model has limited ability to retrieve the arc visual encoding value.}
\label{fig:vlat_pie_chart_problem}
\end{figure}

Previous cases demonstrate the potential limitation of the MLLMs over the Stack Bar Chart. The last error analysis is with the Pie chart.  
With the visualization in Fig.~\ref{fig:vlat_pie_chart_problem} (1), the model responds 17.6\% (4/10) which is the correct answer. However, the rest of the answer is 25.3\% and the detailed decision process of the model can be found in the following response. 

\textit{\textbf{Ask:} “Question: What is the approximate global smartphone market share of Samsung?
Answers: 1) 17.6\%, 2) 25.3\%, 3) 10.9\%, 4) 35.2\% Please answer with Answer: 1) - 4) and Why:”}

\textit{\textbf{Response}: "Answer: 2)
Why: By observing the pie chart, the approximate market share of Samsung is a bit more than a quarter of the total chart. Therefore, it aligns most closely with the option 25.3\%.}

The first hypothesis is that the model missing a value reference to understand the percentage of the area. To verify this, we add a dashed line in the visualization to clear-cut the pie chart into 4 pies, and each represents 25\%.  The new visualization is updated in  Fig.~\ref{fig:vlat_pie_chart_problem} (2). However, the overall response is 4/10 correct answers.
We added additional references in the visualization to identify the region belonging to Samsung, but the final response in Fig.~\ref{fig:vlat_pie_chart_problem} (3) is still 4/10 correct answers.
At the end, we label this region with 17\% on the size. Finally, the model can answer 10/10 that the MLLM correctly responds to the question.

\subsection{Evaluation on VLAT Dataset}

\begin{table*}[t]
\centering
\begin{tabular}{cccccccc}
 \hline
Task/LLM   & GPT4-o & Claude-opus & Gemini-1.5-pro & human & RB\\  \hline
Retrieve Value(13)                                   & 0.461 & 0.338    & 0.361 & \textbf{0.621}  &- \\
Find Extremum (12)                                & 0.683  & 0.491   & 0.675 & \textbf{0.782}&- \\ 
Determine Range (5)                               & 0.5  & 0.48     & \textbf{0.559 } & 0.492   &- \\
Find Correlations/Trends (5)                 & 0.82  & \textbf{0.820}      & 0.840  & 0.774 &-\\ 
Make Comparisons (12)                          & 0.575 & \textbf{0.75}   & 0.525  & 0.648 &- \\ 
Characterize Distribution (1)                 & 0.0      & 0.0     & 0.266 & \textbf{0.457 } & - \\ 
Find Anomalies (2)                                   & 0.35   & 0.2     & 0.2 & \textbf{0.485} &- \\ 
Find Clusters (2)                                        & \textbf{0.85}     & \textbf{0.85}     & 0.05 & 0.751 & - \\ 
identify the hierarchical structure (1)  & \textbf{1.0 }              & \textbf{1.0 }                & 0.9 & 0.879& - \\ 
\hline
All (53) maximum                                               &  0.660                  & 0.622                     & 0.566                &  \textbf{0.943} & -\\ 
All (53) minimum                                                &  \textbf{0.547}                  & 0.452                     & 0.453                 & 0.339 &-\\ 
All (53) mean                               & 0.586     & 0.545      & 0.518 & \textbf{0.667 } & 0.33 \\ 
  \hline
\end{tabular}
\caption{The performance of four MLLMs and humans on the VLAT datasets. }
\label{tab:VLAT_LLM_Task_data_table}
\end{table*}

Given the mostly favorable performance of MLLM over the Mini-VLAT dataset (particular for GPT4-o), we perform the same evaluation over the VLAT dataset with 53 questions to measure the performance of the models on a wider set of tasks and visual encodings.
During the evaluation, one interesting question to explore is whether there is a clear distinguishing pattern between human visualization literacy and MLLM visualization literacy, e.g., where MLLMs align with human behaviors, and where they diverge.

Table~\ref{tab:VLAT_LLM_Task_data_table} provides the summary of the evaluation result over MLLMs and humans over nine tasks. The best performance in each task is highlighted in bold text. RB represents the random baseline, where we random select an option in each question. Both humans and MLLMs perform better than the random baseline.
Comparing the average performance of MLLMs and humans, humans perform better than the MLLMs on average. The maximum performance of a human is much better than that of a MLLM. 
For the minimum performance, MLLM appears to be better than a human. 
However, the reported minimal human performance is close to random selection. Since the answers are crowd-sourced the minimum performance estimation may represent a lack of effort from the participant rather than an accurate estimation of human capabilities.

\begin{figure}[t]
\centering   
    \includegraphics[width=\linewidth]{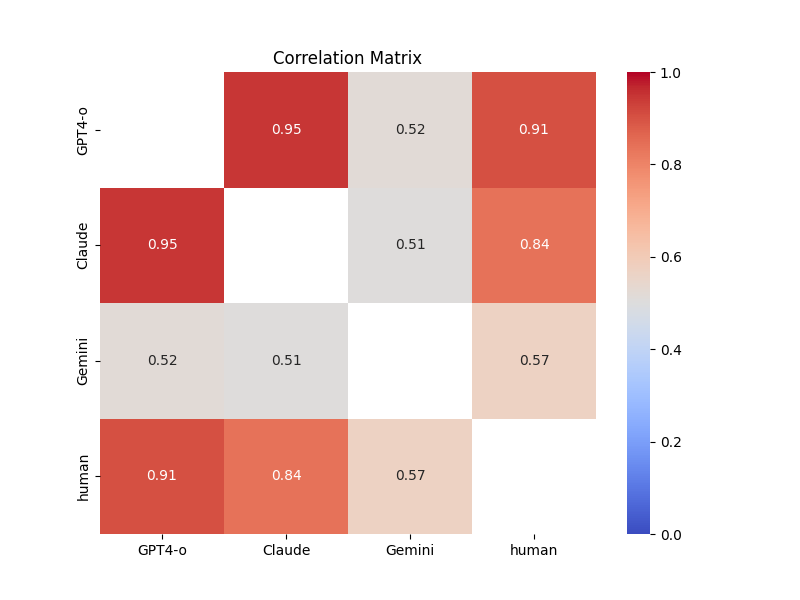}
    \caption{Task-based performance correlation analysis between MLLMs and humans}
\label{fig:vlat_each_task_correlation}
\vspace{-5mm}
\end{figure}

Humans outperform on \textbf{Characterize-Distribution}, \textbf{Retrieve-Value}, and \textbf{Find-Extremum} with a significant margin.
For the rest of the tasks, MLLMs outperform humans.
Especially, MLLMs perform better than humans on tasks such as \textbf{Find Correlations/Trends}, \textbf{Find Clusters}, and \textbf{identify the hierarchical structure}.
If we consider the visualization task as a basic unit to compare the behavior of humans and MLLMs, we can tell from the correlation matrix in Fig.~\ref{fig:vlat_each_task_correlation} that humans' response to visualization has a strong correlation with the MLLMS such as GPT4-o and Claude. Task-based correlation between humans and neural network models is also discovered in previous literature~\cite{giovannangeli2020toward} over graph visualization task.
However, this statement needs more studies to be verified in future studies.
In the meantime, MLLMs also have a strong correlation with each other except Gemini which demonstrates a minor correlation.

Table~\ref{fig:vlat_question_data_table_detail} displays the detailed performance of each question, and its corresponding visual encoding and task type. With 14 over 53 questions, humans outperform MLLMs. However, humans have the best average performance in 6 out of 12 visual encoding types. In the rest of the cases, the MLLM model outperforms humans.
From the evaluation result, we identify some interesting observations.

\begin{figure}[t]
\centering   
    \includegraphics[width=\linewidth]{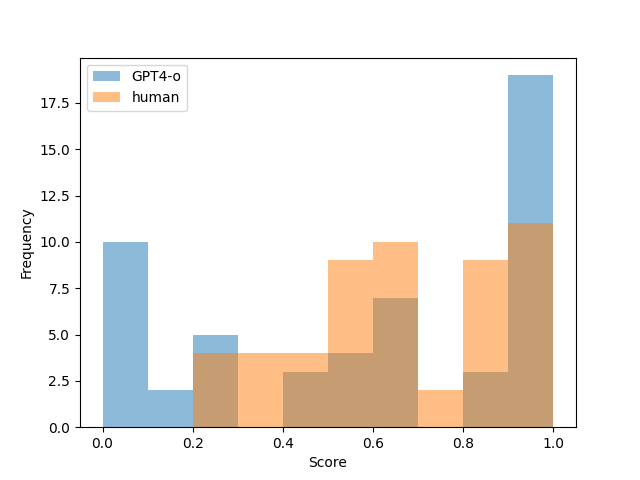}
    \caption{Average score distribution of 53 questions from human and GPT4-o.}
\label{fig:score_distribution_of_human_and_MLLM}
\end{figure}

\begin{figure}[t]
\centering   
    \includegraphics[width=\linewidth]{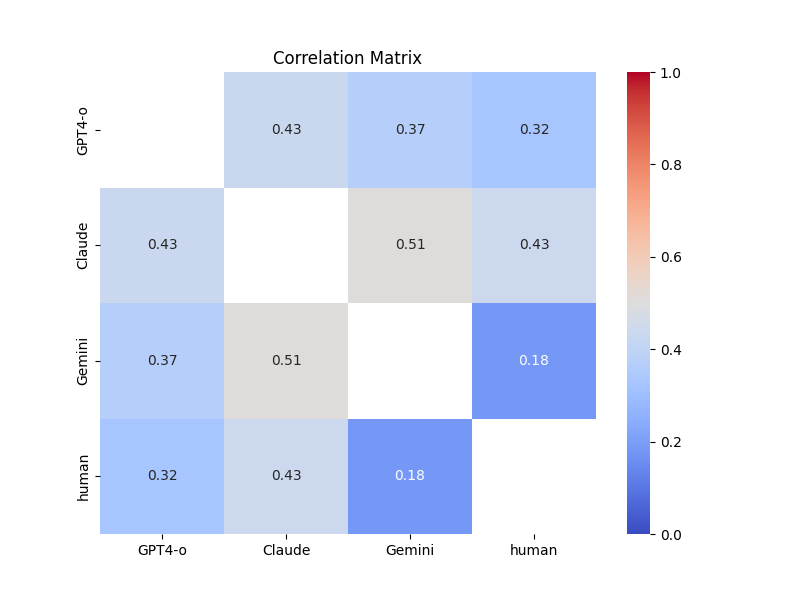}
    \caption{Question-based performance correlation analysis between MLLMs and humans}
\vspace{-5mm}
\label{fig:vlat_each_question_correlation}
\end{figure}

\textbf{Humans have a more stable performance than MLLMs}. The value range of humans' average performance ranges from 0.2 to 1.0, and the overall distribution is coherent.
During the evaluation process, an MLLM may display a 0\% success rate when answering certain questions that do not occur in humans.
On the other hand, MLLMs may show excellent performance in certain tasks such as Line Chart item 2, item 3, and item 4. 
All three models have a 100\% success rate to answer these questions.
From the score distribution of humans and MLLMs in Fig.~\ref{fig:score_distribution_of_human_and_MLLM}, we can conclude that the performance of humans is more stable than MLLMs. A similar observation was also discovered from previous work~\cite{chen2023beyond} with a text-based GPT model.
However, MLLMs have more extreme cases that either perform the test well or badly.
Fig.~\ref{fig:score_distribution_of_human_and_MLLM} just compares GPT4-o with humans but the other two models also show similar properties. 
This observation led to another question whether there is a correlation between humans and MLLMs over question-based scoring.
The answer can be found in Fig.~\ref{fig:vlat_each_question_correlation} that the correlation between humans and MLLMs is not strong. The correlation between MLLMs is also weak.

From the evaluation result of the Mini-VLAT and VLAT datasets, the performance of MLLMs demonstrates a competitive ability in visualization literacy when compared with humans.
We can tell that MLLMs may have certain similarities with humans' behavior over visualization literacy tests but such similarities may not be particularly strong.

\begin{table*}[t]
\centering
\begin{tabular}{ccccccc}
 \hline
Item ID   & Visualization      & Task                     & GPT4-o & Claude-opus & Gemini-1.5-pro & Human \\  
\hline
item1     & {}                 & Retrieve Value                          & 0.2
& 0.0
& 0.0
& \textbf{0.693}\\
item2     & {}                 & Find Extremum                        & \textbf{1.0}& \textbf{1.0}& \textbf{1.0}& 0.964 \\
item3     & Line Chart (1)   & Determine Range               & \textbf{1.0}& \textbf{1.0}& \textbf{1.0}& 0.542 \\
item4     & {}                 & Find Correlations/Trends       & \textbf{1.0}& \textbf{1.0}& \textbf{1.0}& 0.954 \\
item5     & {}                 & Make Comparisons                  & 0.0& 0.5& 0.2& \textbf{0.613}\\
 mean& {}& {}& 0.64& 0.7& 0.64&\textbf{0.753}\\ 
\hline
item1     & {}                 & Retrieve Value                          & 0.4
& 0.3
& 0.1
& \textbf{0.839}\\
item2     & Bar Chart(2)   & Find Extremum                   & \textbf{1.0}& 0.8
& \textbf{1.0}& 0.959 \\
item3     & {}                 & Determine Range                    & 0.0
& 0.3
& 0.0
& \textbf{0.593}\\
item4     & {}                 & Make Comparisons                 & 0.4& 0.5& \textbf{0.8}& 0.261 \\
 mean& {}& {}& 0.449& 0.475& 0.475&\textbf{0.663}\\ 
\hline
item1     & {}                 & Retrieve Value                         & 0.2
&        0.0
&                0.0
& \textbf{0.884}\\
item2     & {}                 & Retrieve Value                         & 0.5
&        \textbf{0.6}&                0.1
& 0.402 \\
item3     & Stacked Bar Chart(3)  & Find Extremum     & 0.0
&        0.6
&                0.0
& \textbf{0.703}\\
item5     & {}                 & Make Comparisons                & 0.0
&        \textbf{1.0}&                0.0
& 0.568 \\
item6     & {}                 & Make Comparisons                & 0.4&        0.7&                \textbf{0.9}& 0.557 \\
 mean& {}& {}& 0.22& 0.58& 0.2&\textbf{0.623}\\
\hline
item1     & {}                 & Retrieve Value                                       & \textbf{1.0}&        0.8
&                \textbf{1.0}& 0.507 \\
item2     & 100\% Stacked Bar Chart(4)  & Find Extremum        & 0.0
&        0.0
&                0.5
& \textbf{0.819}\\
item3     & {}                 & Make Comparisons                              & 0.3&        \textbf{1.0}&                0.8& 0.583 \\
 mean& {}& {}& 0.433& 0.6& \textbf{0.766}&0.636\\
\hline
item1     & {}                 & Retrieve Value                     & 0.8
&        \textbf{1.0}&                \textbf{1.0}& 0.693 \\
item2     & Pie Chart(5)      & Find Extremum             & \textbf{1.0}&        \textbf{1.0}&                0.8
& 0.979\\
item3     & {}                 & Make Comparisons             & \textbf{1.0}&        \textbf{1.0}&                0.8& 0.944 \\
 mean& {}& {}& 0.933& \textbf{1}& 0.866&0.872\\
\hline
item1     & {}                 & Retrieve Value                    & 0.7
&        \textbf{1.0}&                0.2
& 0.914 \\
item2     & Histogram(6)   & Find Extremum            & 0.8
&        0.2
&                \textbf{1.0}& 0.934 \\
item4     & {}                 & Make Comparisons           & 0.0&        \textbf{0.9}&                0.8& 0.879 \\
 mean& {}& {}& 0.5& 0.7& 0.666&\textbf{0.909}\\
\hline
item1     & {}                 & Retrieve Value                      & 0.0
&        0.1
&                0.2
& \textbf{0.633}\\
item2     & {}                 & Find Extremum                    & 0.1
&        0.6
&                \textbf{0.9}& 0.633 \\
item3     & {}                 & Determine Range                 & 0.6
&        0.5
&                \textbf{1.0}& 0.527 \\
item4     & Scatterplot(7)  & Find Anomalies              & 0.0
&        0.0
&                \textbf{0.4}& 0.341 \\
item5     & {}                 & Find Clusters                        & 0.7
&        0.8
&                0.0
& \textbf{0.854}\\
item6     & {}                 & Find Correlations/Trends  & \textbf{1.0}&        \textbf{1.0}&                0.9
& 0.683 \\
item7     & {}                 & Make Comparisons             & \textbf{1.0}&        \textbf{1.0}&                \textbf{1.0}& 0.778 \\
 mean& {}& {}& 0.485& 0.571& 0.628&\textbf{0.635}\\
\hline
item1     & {}                 & Retrieve Value                             & 0.5
&        0.0
&                0.0
& \textbf{0.683}\\
item2     & Area Chart(8)  & Find Extremum                     & 0.6
&        0.6
&                \textbf{0.8}& 0.396 \\
item3     & {}                 & Determine Range                       & 0.2
&        0.6
&                \textbf{0.8}& 0.381 \\
item4     & {}                 & Find Correlations/Trends         & \textbf{1.0}&        0.9&                \textbf{1.0}& 0.944 \\
 mean& {}& {}& 0.575& 0.525& \textbf{0.65}&0.601\\
\hline
item1     & {}                 & Retrieve Value                           & 0.7
&        0.1
&                \textbf{0.9}& 0.201 \\
item2     & {}                & Retrieve Value                            & \textbf{0.7}&        0.0
&                0.2
& 0.296 \\
item3     & Stacked Area Chart(9)  & Find Extremum     & \textbf{1.0}&        0.0
&                0.1
& 0.909\\
item4     & {}                 & Find Correlations/Trends        & \textbf{0.7}&        0.0
&                0.0
& 0.417 \\
item5     & {}                 & Make Comparisons                   & \textbf{1.0}&        \textbf{1.0}&                \textbf{1.0}&0.959 \\
item6     & {}                 & Make Comparisons                   & \textbf{0.8}&        0.5&                0.5& 0.286 \\
 mean& {}& {}& \textbf{0.816}& 0.266& 0.45&0.511\\
\hline
item1     & {}                 & Retrieve Value                          & 0.0
&        0.4
&                \textbf{1.0}& 0.864 \\
item2     & {}                 & Find Extremum                       & \textbf{0.7}&        0.1
&                0.0
& 0.648 \\
item3     & {}                 & Characterize Distribution                     & 0.0
&        0.0
&                0.0
& \textbf{0.457}\\
item4     & Bubble Chart(10)        & Find Anomalies      & \textbf{0.7}&        0.4
&                0.0
&0.628 \\
item5     & {}                 & Find Clusters                            & \textbf{1.0}&        0.9
&                0.1
& 0.648 \\
item6     & {}                 & Find Correlations/Trends      & 0.1
&        0.2
&                0.3
& \textbf{0.331}\\
item7     & {}                 & Make Comparisons                 & \textbf{1.0}&        0.2&                0.0& 0.532 \\
 mean& {}& {}& 0.5& 0.314& 0.2&\textbf{0.586}\\
\hline
item1     & {}                 & Retrieve Value                                       & 0.3
&        0.1
&                0.0
& \textbf{0.467}\\
item2     & Choropleth Map(11)          & Find Extremum            & \textbf{1.0}&        0.0
&                \textbf{1.0}&0.507 \\
item3     & {}                 & Make Comparisons                               & \textbf{1.0}&        0.8&                0.0& 0.889\\
 mean& {}& {}& \textbf{0.766}& 0.3& 0.33&0.621\\
\hline
item1     & {}                 & Retrieve Value                                       & \textbf{1.0}&        \textbf{1.0}&                \textbf{1.0}& 0.934\\
item2     & Treemap(12)         & Make Comparisons                     & \textbf{1.0}&        0.9
&                0.5
& 0.879 \\
item3     & {}                 & Identify the Hierarchical Structure    & \textbf{1.0}&        \textbf{1.0}&                0.9& 0.879 \\
 mean& {}& {}& \textbf{1}& 0.967& 0.799&0.897\\
 \hline

\end{tabular}
\caption{The table displays the performance of LLM and humans over the VLAT datasets. The table highlights the best performance in each question.}
\label{fig:vlat_question_data_table_detail}
\end{table*}

\section{Discussion and Conclusion}
The new emerging visual perception ability of the MLLMs brings new opportunities and challenges to the visualization community.
How this new ability will change the way humans interact with visualization is an exciting research direction in the expected future.
A critical step to utilize this new technique is to rely on the proper evaluation of the MLLMs to reveal their advantages and limitations.
In this study, we leverage the Mini-VLAT and VLAT datasets, the popular dataset for visualization literacy measurement, to understand MLLMs' capability to accomplish basic visualization understanding tasks.
Our evaluation tells us that MLLMs already demonstrate competitive performance over these visualization literacy tests, and can perform certain tasks such as correlation/trend analysis, and treemap analysis very well. MLLMs and humans demonstrate different behavior patterns in their question score distribution and the question-based correlation may not be particularly strong. 
Our error analysis with the MLLMs also reveals that these models still have quite a few limitations for the reliable interpretation of visualization and visual encodings.
The MLLMs still struggle to interpret the basic elements such as the portion of a pie chart with additional annotation in the visualization.

Our study mainly focuses on the basic prompt to understand the performance of the multi-model large language model.
Prompt engineering can play a significant role in the performance of the model. 
It would be interesting to explore how different prompting techniques \cite{wei2022chain} for MLLMs can affect the overall performance, and whether they are effective for vision tasks. 
We believe our study represents one of the necessary steps for revealing the truth capability of MLLM for visualization interpretation.
There is a broad spectrum of research that needs to be done to evaluate the ability of MLLMs such as how MLLMs react to different visual channels~\cite{mccoleman2021rethinking,heer2010crowdsourcing}, and how MLLMs react to more advanced visualization techniques such as parallel coordinate~\cite{inselberg1990parallel} visualization.


\acknowledgments{
This work was performed under the auspices of the U.S. Department of Energy by Lawrence Livermore National Laboratory under Contract DE-AC52-07NA27344. This work is partially supported by LDRD 23-ERD-029. The manuscript is reviewed and released under LLNL-PROC-865825.
}

\clearpage
\bibliographystyle{abbrv-doi}
\bibliography{lib}

\end{document}